# PECNet: A Deep Multi-Label Segmentation Network for Eosinophilic Esophagitis Biopsy Diagnostics

Nati Daniel[1,#], Ariel Larey[2,#], Eliel Aknin[2,#], Garrett A. Osswald[3,#], Julie M. Caldwell[3], Mark Rochman[3], Margaret H. Collins[4], Guang-Yu Yang[5], Nicoleta C. Arva[6], Kelley E. Capocelli[7], Marc E. Rothenberg[3], Yonatan Savir[1,*]

[1]Dept. of Physiology, Biophysics and System Biology, Faculty of Medicine, Technion, 1st Efron st. Haifa, 35254, Israel.
[2]Faculty of Electrical Engineering, Technion – Israel Institute of Technology, Haifa 3200003, Israel.
[3]Division of Allergy and Immunology, Cincinnati Children's Hospital Medical Center, Department of Pediatrics, University of Cincinnati College of Medicine, Cincinnati, OH 45229-3026, USA.
[4]Division of Pathology, Cincinnati Children's Hospital Medical Center, Department of Pediatrics, University of Cincinnati College of Medicine, Cincinnati, OH 45229-3026, USA.
[5]Division of Pathology, Northwestern University, The Feinberg School of Medicine
[6]Pathology Department, Ann & Robert H. Lurie Children's Hospital of Chicago, Northwestern University Feinberg School of Medicine, Chicago, IL, USA.
[7]Division of Pathology, Children's Hospital Colorado, Aurora, CO.

[#]These authors contributed equally to this work
[*]Correspondence: yoni.savir@technion.ac.il

## Summary

**Background** Eosinophilic esophagitis (EoE) is an allergic inflammatory condition of the esophagus associated with elevated numbers of eosinophils. Disease diagnosis and monitoring requires determining the concentration of eosinophils in esophageal biopsies, a time-consuming, tedious and somewhat subjective task currently performed by pathologists.

**Methods** Herein, we aimed to use machine learning to identify, quantitate and diagnose EoE. We labeled more than 100M pixels of 4345 images obtained by scanning whole slides of H&E-stained sections of esophageal biopsies derived from 23 EoE patients. We used this dataset to train a multi-label segmentation deep network. To validate the network, we examined a replication cohort of 1089 whole slide images from 419 patients derived from multiple institutions.

**Findings** PECNet segmented both intact and not-intact eosinophils with a mean intersection over union (mIoU) of 0.93. This segmentation was able to quantitate intact eosinophils with a mean absolute error of 0.611 eosinophils and classify EoE disease activity with an accuracy of 98.5%. Using whole slide images from the validation cohort, PECNet achieved an accuracy of 94.8%, sensitivity of 94.3%, and specificity of 95.14% in reporting EoE disease activity.

**Interpretation** We have developed a deep learning multi-label semantic segmentation network that successfully addresses two of the main challenges in EoE diagnostics and digital pathology, the need to detect several types of small features simultaneously and the ability to analyze whole slides efficiently. Our results pave the way for an automated diagnosis of EoE and can be utilized for other conditions with similar challenges.

## Introduction

Eosinophilic esophagitis (EoE) is a chronic allergic inflammatory condition of the esophagus characterized by elevated levels of esophageal eosinophils[1]. After gastroesophageal reflux disease, EoE is the most common cause of chronic esophagitis leading to symptoms such as dysphagia and esophageal food impaction[2]. Food hypersensitivity, allergic inflammation, and multiple genetic and environmental factors are the main drivers of the disease pathogenesis[3]. The diagnosis of EoE requires manual microscopic review of endoscopic biopsies, and the diagnostic threshold of at least 15 eosinophils/400X high-power field (HPF) is required.

Hematoxylin and eosin (H&E) staining is used frequently to detect eosinophilic cells, as their unique basically charged granule constituents have affinity for the eosin stain[4]. A common practice is to identify the area of tissue within a slide that exhibits the most dense eosinophil infiltration and quantify the peak eosinophil count (PEC) in that particular HPF and compare it to a predetermined threshold[4–6]. It is crucial to distinguish between intact eosinophils, which have both their intensely red cytoplasmic granules and nucleus visible[6] and contribute to the PEC, and not-intact eosinophils without granules or visible nucleus that are not added to the PEC. Detecting and counting different cellular features is a laborious and time-consuming task that leads to inconsistencies even between trained observers[4–6]. The field still lacks a robust automatic process that can cope with the task of counting inflammatory cells such as eosinophils and aid the pathologists.

Deep learning can be utilized to develop methods to complete various tasks in digital pathology[7]. With the approach of semantic segmentation, a tissue image is labeled to designate local regions of the image as belonging to certain classes, and the network is then trained to identify local regions belonging to the same classes in separate images. Previously, this approach has been applied successfully to identify types of cancerous lesions[8,9], to segment cell nuclei[10], to segment inflammatory bowel disease (IBD) tissue features[11], to classify different cancer types via histology images[12,13], to perform cancer screening[14], to personalize cancer care[15], and to perform fundamental deep learning methods in medical field analysis[16], such as image registration, detection of anatomical and cellular structures, tissue segmentation, and computer-aided disease diagnosis and prognosis. Various net architectures have been developed for segmentation, such as Mask RCNN[17], Mask ECNN[18], Deep Lab[19,20], and Generative Adversarial Networks[21]. A common architecture model for biomedical image segmentation, coined Unet[22], is based on auto-encoder architecture in which the encoder takes the input, performs down-sampling and outputs a feature vector/map that represents the input, and the decoder does the opposite orientation (i.e. up-sampling) that takes the feature vector (i.e. the features) from the encoder, and gives the best closest match to the actual input or intended output. An upgraded model of the Unet, called Unet++[23], introduced re-designed skip pathways between the original Unet layers. These pathways aim to reduce the semantic gap between the feature maps of the encoder and decoder sub-networks. Another update for Unet++ is the deep supervision that combines different sub-models of the full Unet++ to segment the image and enables more accurate segmentation, particularly for lesions that appear at multiple scales.

A second approach is to provide a label to an entire slide according to one or more conditions (e.g., whether a patient has active EoE or not) and to train the network without local labeling of the image. The main advantage of this approach is that it does not require intensive local labeling effort and allows the machine to infer representations of a condition without a priori local bias. We have recently developed a deep learning system for classifying H&E-stained esophageal tissue images based on global labeling[24]. Using this approach led to a classification of EoE disease activity with an accuracy of approximately 85% based only on global labels. Furthermore, these results showed that histological features associated with EoE are not only local clusters of eosinophils but also global attributes of the histology pattern.

Here, we develop a multi-label semantic segmentation approach based on Unet++ that processes whole slide images and segments both intact and not-intact eosinophils. Our net has excellent segmentation performances with a mean intersection over union metric (mIoU) score of 0.93 that allows counting eosinophils with a standard error of 1.68. Based on eosinophil counts, the network is able to classify EoE disease activity according to whether their PEC is greater than 15/HPF with an accuracy of 98.5%, sensitivity of 96.9%, and specificity of 98.9%. Finally, we validated our network on a cohort of 1089 biopsy slides from 419 patients and achieved an accuracy of 94.8%, sensitivity of 94.3%, and specificity of 95.1%.

## Methods

### Study population and datasets

This study was conducted within the context of the Consortium of Eosinophilic Gastrointestinal Disease Researchers (CEGIR)[25], a national collaborative network of 16 academic centers caring for adults and children with eosinophilic gastrointestinal disorders. The CEGIR clinical trial, Outcomes Measures in Eosinophilic Gastrointestinal Disorders across the Ages (OMEGA), is a longitudinal cohort study aimed at understanding the natural history of EoE, eosinophilic gastritis, and eosinophilic colitis during routine clinical care. All subjects' clinical data were stored at the Data Management and Coordinating Center (DMCC) at Cincinnati Children's Hospital Medical Center. Data were systematically extracted from the databases. This study was approved by the institutional review boards of the participating institutions via a central institutional review board at Cincinnati Children's Hospital Medical Center (CCHMC IRB protocol 2015-3613). Participants provided written informed consent. 419 subjects undergoing endoscopy (EGD) for standard-of-care purposes agreed to have their clinical, histologic, and demographic information stored in a private research database. Distal, mid, or proximal esophageal biopsies (1-3 per anatomical site) per patient were placed in 10% formalin; the tissue was then processed and embedded in paraffin. Sections (4 μm) were mounted on glass slides and subjected to hematoxylin and eosin (H&E) staining. Slides were scanned on the Aperio scanner at 400X magnification and were saved in .SVS format. Each slide of esophageal tissue was analyzed by an anatomic pathologist who is a member of the CEGIR central pathology core to determine peak eosinophil count per 400X high-power field (HPF). The peak eosinophil counts associated with each image were stored at the DMCC.

### Semantic labeling

WSI images from 23 patients were chosen for semantic labeling. These images were split into patches with a size of 1200X1200 pixels. Patches (n = 10,736) contained less than 85% background and were annotated by a trained experienced researcher. For each patch, each pixel was assigned to one of three classes: intact eosinophils, defined by eosinophils with visible intensely red cytoplasmic granules and nucleus; not intact eosinophils, defined by eosinophils without a visible nucleus or large groups of extracellular eosinophil granules (Table 1); or non-eosinophils.

## PECNet training procedure

The labeled images were split arbitrarily into training (80%) and validation (20%) sets. Using a rectangular grid, each image was cropped into nine 448X448 pixel patches. These patches were the input into the network based on the implementation of Unet++. The updated model was trained and optimized using Pytorch[26] framework on a single NVIDIA GeForce RTX 2080 Ti GPU. During the training, different hyperparameters were examined. The "Cosine Annealing" learning rate, 448X448 patch size, batch size of 5, 100 epochs, and 0.5 Softmax Threshold were revealed to be optimal. Moreover, the optimal loss function includes a Dice metric and a binary cross-entropy (BCE) element, where the Dice and BCE are weighted with values of 1 and 0.5, respectively. The output of the model was converted into a binary segmentation mask using a 0.5 probability threshold. The masks of the different patches were combined to reconstitute the original 1200X1200 image using OR function (i.e. is the truth-functional operator of inclusive disjunction) for the overlapping regions.

| Class | Number of images (percent of total)[a] | Total Area (pixels)[b] | Mean fraction out of total patch area[c] |
|---|---|---|---|
| **Eos-not intact** | 2028 (18.89%) | 24.37M | 0.83% |
| **Eos-intact** | 2317 (21.58%) | 78.47M | 2.35% |

**Table 1:** Semantic labeling statistics at the 1200X1200 patches level. Whole slide images (WSIs) from 23 subjects were split into 1200X1200 pixel patches. The patches with <85% background (n = 10,736) were annotated to identify all eos-intact and eos-not intact within each image; the remainder of the image area was by default classified as non-eosinophil.

[a], The total number (with percentage out of 10,736 total patches indicated in parentheses) of 1200X1200 pixel images containing at least one instance of the indicated class of eosinophil.

[b], The sum of the area (in pixels) of the indicated class of eosinophil from the 10,736 images.

[c], For each 1200X1200 pixel patch, the number of pixels denoted as the indicated class of eosinophil was divided by the total number of pixels; the average percentage of pixels classified as the indicated class per total patch area for the number of images reported in column [a] was then calculated.

Eos, eosinophils; M, million.

## Segmentation metrics

To estimate the segmentation performances, we used the following metrics,

$$mIOU = \frac{1}{I \cdot C} \sum_i \sum_c \frac{TP_{i,c}}{TP_{i,c} + FP_{i,c} + FN_{i,c}}$$ (1)

$$mPrecision = \frac{1}{I \cdot C} \sum_i \sum_c \frac{TP_{i,c}}{TP_{i,c} + FP_{i,c}}$$ (2)

$$mRecall = \frac{1}{I \cdot C} \sum_i \sum_c \frac{TP_{i,c}}{TP_{i,c} + FN_{i,c}}$$ (3)

$$(4) \quad mSpecification = \frac{1}{I \cdot C} \sum_i \sum_c \frac{TN_{i,c}}{TN_{i,c} + FP_{i,c}}$$

where the $c$ index iterates over the different classes in the image, and the $i$ index iterates over the different images in the dataset. $C$ is the total number of classes, and $I$ is the total number of images. TP, TN, FP, FN are classification elements that denote true positive, true negative, false positive, and false negative of the areas of each image, respectively.

**Estimating the PEC**

At the dataset resolution, the average area of a typical eosinophil is 2050 pixels. Each connected region with an area larger than 1800 and less than or equal to 3000 pixels was counted as one eosinophil. In a connected region that is larger than 3000 pixels, each additional 2000 pixels adds one eosinophil to the count of this connected region. The classification of EoE disease activity of an image was based on calculating the number of intact eosinophils per 400X HPF (area of 0.3mm$^2$) being greater than or equal to 15 (active EoE), or less than 15 (inactive EoE or non-EoE).

**PECNet Architecture**

Our network's input can be either one EoE patient's whole slide image (WSI) or several patients' WSIs, whereas the output includes 3 elements: 1. Segmented WSIs – each pixel of the WSIs is labeled as belonging to the relevant class (i.e., intact eosinophils, not intact eosinophils, or non-eosinophils). This output would aid the pathologist's search for important regions. 2. PEC, which represents the intact intraepithelial eosinophil count of the most densely infiltrated high-power field for each WSI. 3. A ranked eosinophil count list, which can help the pathologist prioritize between severe to non-severe patients. The overall flow is described in Figure 1. For inference, first, the WSI is cropped into patches to be adjusted to the deep learning net input size. Second, every patch is fed into the model network and segmented. Finally, all the segmentation patches masks are fused back to the full size of the original WSI, and a full segmented EoE patient's WSI is produced. Based on this output, the WSI's global PEC is calculated by counting intact eosinophils using a kernel in the size of the HPF of an area of 0.3mm$^2$ and iterating over the segmented slide to find the maximum eosinophil count within the HPF boundaries.

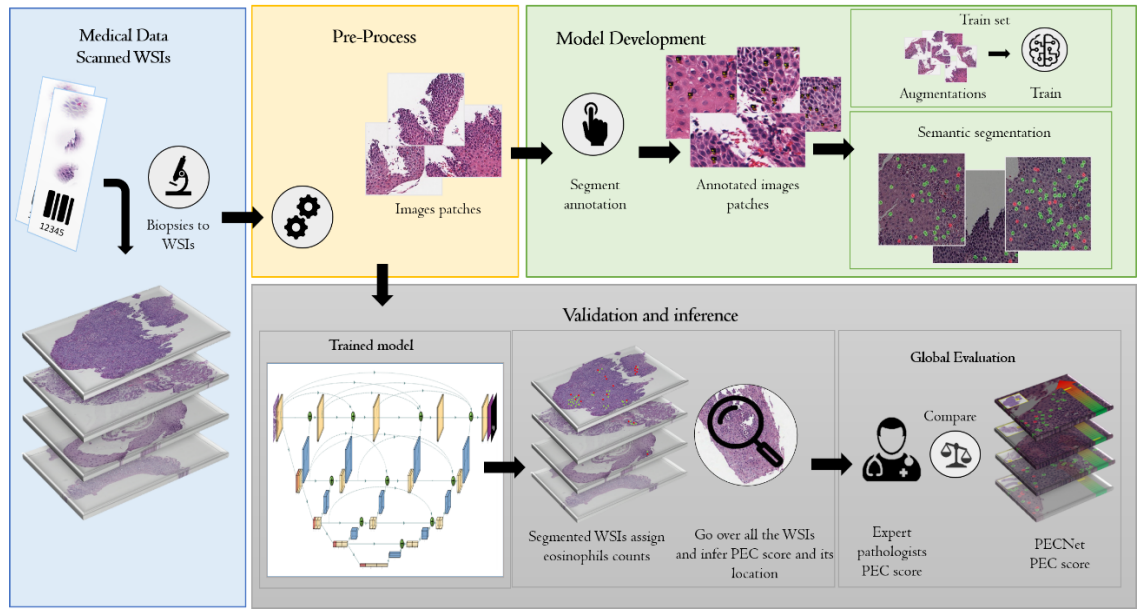

**Figure 1: Overview of PECNet's training and procedure.** PECNet high-level architecture is presented. For training, whole slide images (WSIs) were pre-processed and cropped into patches on which a trained researcher marked the location of both intact and not intact eosinophils. The trained multi-label semantic segmentation network is able to get a WSI as an input, segment it, count the two types of eosinophils, and locate the area with the highest number of eosinophils, count its value, and calculate the peak eosinophil count per high power field (PEC score). PECNet was validated using 1089 whole image slides from 419 patients for which the PEC had been previously determined by anatomic pathologists.

## Results

**Segmentation**

We first validated the ability of the network to detect and segment both intact and not-intact eosinophils. The network detects both intact and not-intact eosinophils with mIOU (that is, the intersection between the actual and predicted areas divided by their union) of 93%, mPrecision (that is, the fraction of the true positive out of the total predicated area) of 95% for intact eosinophils and 97% for not intact eosinophils, and mRecall (that is the fraction of the true positive out of the total ground truth area) of 97% for intact eosinophils and 95% for not-intact eosinophils (Table 2).

|  | **Intact eosinophils** | **Not-intact eosinophils** | **Overall** |
|---|---|---|---|
| **mIOU** | 0.93 | 0.93 | 0.93 |
| **mPrecision** | 0.95 | 0.97 | 0.96 |
| **mRecall** | 0.97 | 0.95 | 0.96 |
| **mSpecificity** | 0.998 | 0.999 | 0.9988 |

**Table 2:** Segmentation metrics scores on the validation set. Validation cohort whole slide images (n = 1089 images from n = 419 subjects) were input into the network. The ratios for the given metrics are shown for each individual category of eosinophil as well as for the sum of both categories (overall). Eos, eosinophil; IoU, intersection over union; m, mean.

We also evaluated the ability of the network to distinguish between the two types of eosinophils. We determined whether pixels that had a true label of eosinophils (whether it is intact or not-intact) were predicted to be eosinophils and assigned to the correct class. The confusion matrix is shown in Figure 2. Out of the true intact eosinophils that were identified as eosinophils, 98.8% were identified correctly as intact. This performance is critical to the ability of the network to provide a reliable intact eosinophil count.

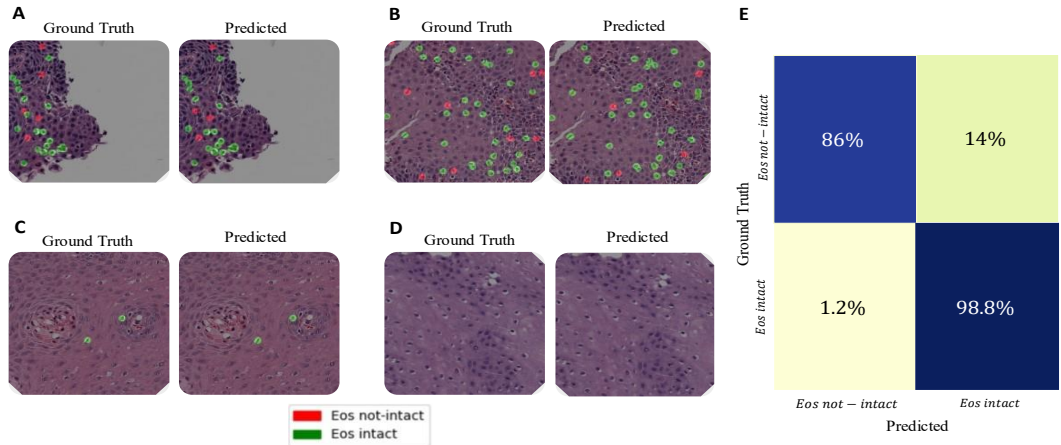

**Figure 2: Quality of multi-label segmentation.** (A-D) Examples of multi-label segmentation. The size of each patch is 1200X1200 pixels. The left images are colored with the ground truth mask's features, and the right replication is colored with its corresponding prediction mask's features. The intact Eos class is colored in green, and the not-intact Eos class is colored in red. (A) and (B) are examples of slides from EoE patients with many eosinophils per patch/HPF, (C) is an example of an EoE patient with only two eosinophils at the local level per patch, and (D) is an example of an EoE patient without any eosinophils at the local level per patch. (E) The confusion matrix between intact and not-intact Eos classes.

In the next stage, we used the segmentation masks to infer the number of intact eosinophils (Methods) and compared their number to the actual semantic labeling. Figure 3 illustrates the relationship between the actual counts and the predicted ones. The best linear fit for the relationship between the predicted counts and ground truth has a slope of 1.005 (95% confidence interval values of 0.963 and 1.013) and an intersection of 0.03 (confidence interval values of -0.065 and 0.126), R-square of 0.9721 and F-test p-value << 0.001 (Fig. 3A). That is, we observed no inherent bias in the counting estimation. The mean count error is 0.611 eosinophils. Another measure for the quality of the segmentation and counting is that both the counting error and the actual false discovery rate of the labeled masks are low and correlated (Fig. 3B). In 90% of the images, the connected segment's false discovery is smaller than 0.1, and the counting error rate is lower than 3%. As such, the correct counting is indeed the result of correct local detection. Next, we estimated the maximal eosinophil count within a patch. The results yield a classifier with an accuracy of 98.48%, sensitivity of 96.89%, and specificity of 98.89% (Fig. 3C).

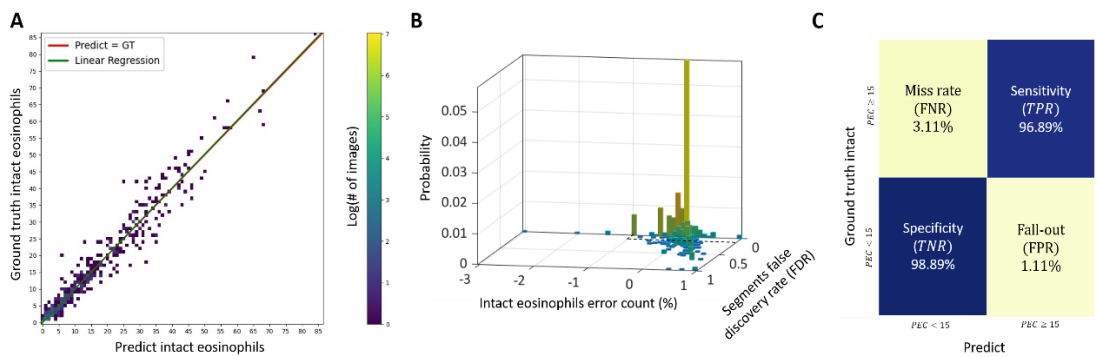

**Figure 3: Quality of intact eosinophil counting.** (A) The relationship between the true number of eosinophils and the number estimated from semantic segmentation of a 1200X1200 patch. The best linear fit has a slope of 1.005 (95% confidence interval values of 0.963 and 1.013), an intersection of 0.03 (confidence interval values of -0.065 and 0.126), and a R-square of 0.9721. The mean absolute counting error is 0.611 eosinophils. (B) In 90% of the images, both the connected segments' false discovery is smaller than 0.1, and the counting error rate is lower than 3% (the area defined by the black dashed lines). (C) Classification confusion matrix based on comparing the peak eosinophil count in a high-power field to 15.

**PECNet Validation**

To validate the ability of PECNet to estimate the peak eosinophil count of a whole slide, we examined a cohort of 1089 biopsy slides from 419 patients. In this cohort, each WSI was derived from a scan done at 400X magnification, resulting in input images with a width of 20K-100K pixels and a typical length of 100K-200K pixels (Fig. 4A). These slides were scored by an anatomic pathologist who examined the slides and estimated the PEC per slide, that is, the maximal intraepithelial eosinophil count observed in an HPF of $0.3mm^2$, which corresponds to a size of 2144 X 2144 pixels. PECNet took a WSI as an input, broke it into 10K-100K patches of 448 X 448 pixels, depending on the original WSI size, segmented each patch, and estimated the number of eosinophils per patch. At the same time, PECNet merges the patches and locates the position and count of the HPF with the maximal eosinophil count (Fig. 4). This procedure was done using dynamic convolution, which allows PECNet to run rapidly within the processing speed and memory constraints of a standard computer.

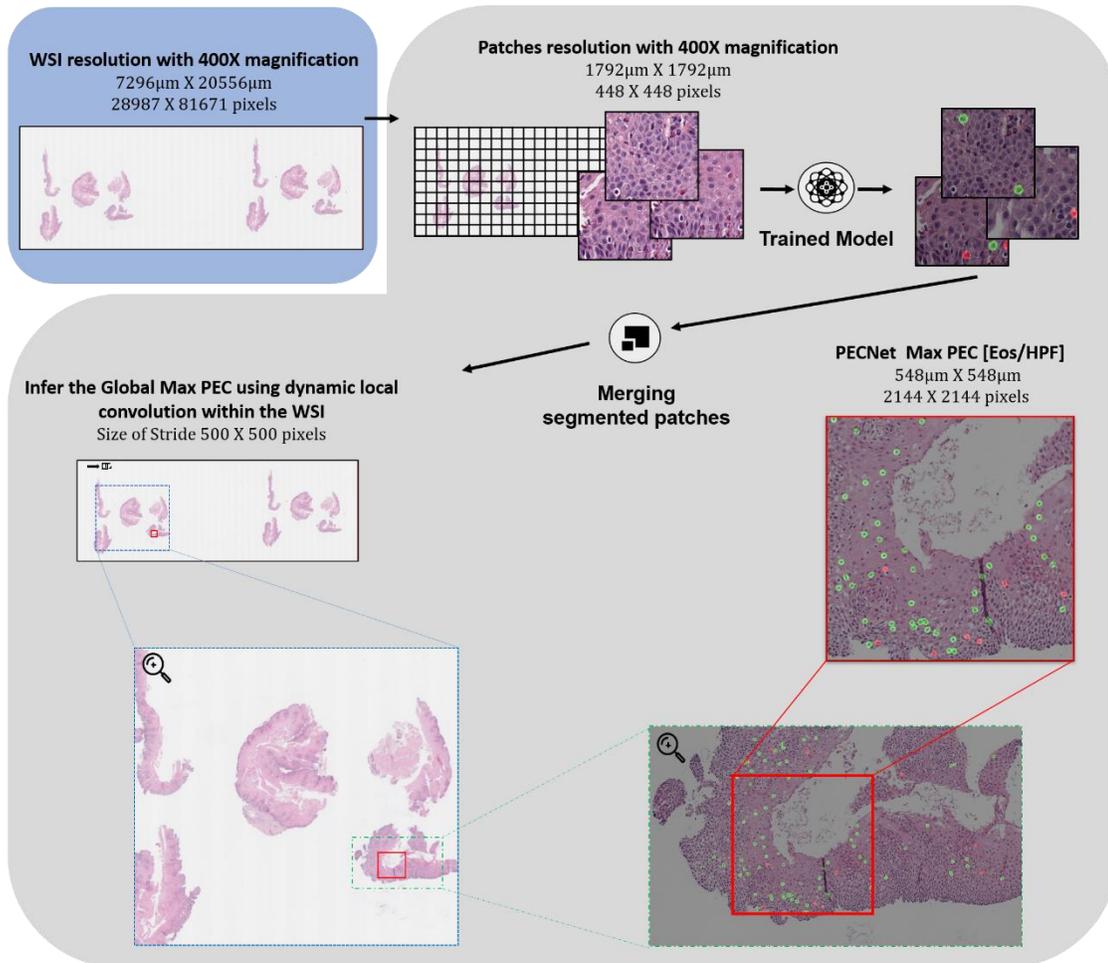

**Figure 4: Applying PECNet to whole slide images (WSIs).** The figure presents an example of an analysis of a specific WSI in the different scales. The red box denotes the location of the HPF where PECNet's maximum PEC was found.

The clinical diagnostic threshold for classifying patients with active EoE is a count of greater than or equal to 15 eosinophils in at least one HPF. Figure 5 shows the comparison between the PECNet count and the anatomic pathologist's count, and the resulting classification. Note that in this case, the algorithm identifies the HPF with the highest number of eosinophils without prior knowledge of the specific HPF chosen by the pathologist. Taking 15 eosinophils as a threshold, the net provides an accuracy of 94.8% with a sensitivity of 94.31% and specificity of 95.14% (Fig. 5A, 5B). The area under the curve is 98.81% and the critical threshold that maximizes the accuracy is 14.5 (Fig. 5C). The classification accuracy for defining PEC in the biopsies obtained from proximal, mid and distal esophagus were 96.2%, 92.9%, and 93.3%, respectively.

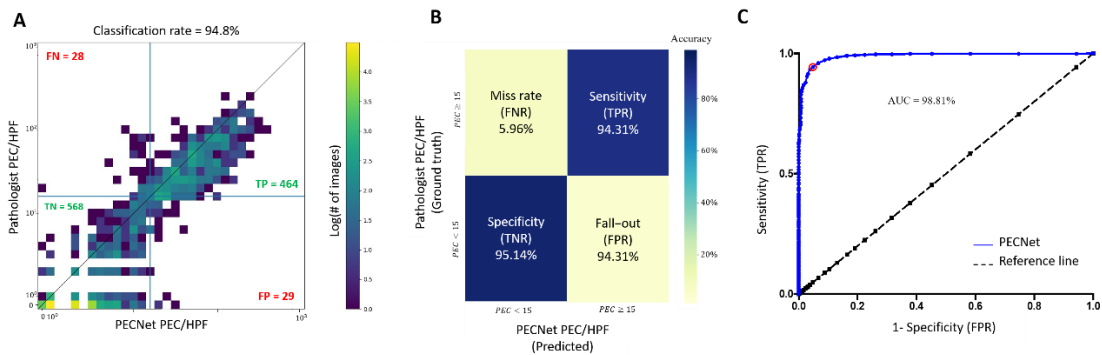

**Figure 5: PECNet validation on a cohort of 1089 slides from 419 EoE patients.** (A) Comparing the maximal intact eosinophils per HPF ($0.3mm^2$) of an entire slide, as measured by the pathologists and PECNet. (B) Classification confusion matrix based on comparing the peak eosinophil count in a high-power field to 15. (C) The classifier performance in the AUC space; each dot represents a different PEC threshold. A threshold of 14.5 (red circle) maximize the overall accuracy.

## Discussion

The digital transformation of pathology is expected to grow dramatically over the next few years as increasing numbers of laboratories gain access to high throughput scanning devices and aim to automate analysis of scanned microscopic images. This transformation is driven by several factors including the prolonged time per case due to the growing complexity of pathological criteria for diagnosing and monitoring diseases compounded by limited supply of pathologists, especially in different geographical regions.

There are inherent challenges in digital pathology beyond data collection. One of the main challenges is the large textual variation of slides and the existence of multiple length scales – slides that are very large compared with the features that define the clinical condition. Moreover, the typical size of a slide is much larger than the typical image input size of current convolutional networks. Thus, even if a label exists at the slide level, training is challenging. On the other hand, labeling at the pixel level is laborious, in particular, if there are multiple tissue features that are relevant for the clinical condition. Herein, we developed a semantic segmentation approach that is trained to identify two classes of eosinophils simultaneously. This approach is different from training two different networks to identify each individually and relieve the problem of conflicting overlap regions.

EoE is an example of a condition that relies on the identification and counting of small objects, such as single cells, within a whole slide. The pathologist is required to scan the slide and evaluate the local concentration of eosinophils. The typical region that is used for counting is the size of 0.3mm$^2$, and thus it is laborious to probe the entire slide. The validation cohort we used in the study included 1089 biopsy slides from 419 patients. Each slide has a PEC score of the maximal eosinophil density as determined by a body of expert pathologists associated with CEGIR. This score is based on the pathologist's conclusion and reflects only the number of high-power fields that the pathologist probed. Thus, a clear advantage to the PECNet is that every possible high-power field can be probed for each slide. To generate a training dataset with exact local information, patches from 23 slides independent of the validation cohort were annotated at the pixel level. We generated a dataset based on greater than 100M labeled pixels from eosinophils. The network provides state of the art segmentation of eosinophils that can discriminate between intact and not-intact eosinophils (Table 2). This segmentation results in a intact eosinophil count that differs from the researcher counts by a mean absolute error of 0.611 eosinophils on identical slides. Taking a cutoff of 15 counts per 0.3mm$^2$ value, the counting on the annotated dataset gives classification accuracy of 98.48%.

While the network has excellent performances on the patch level, each patch's area is three orders of magnitude smaller than the area of a whole slide image. PECNet uses a dynamics convolution approach to segment the entire slide rapidly and estimates the maximal eosinophil density of any given high power field size. This is an important feature as different pathologists may use distinct sizes of a HPF, leading to ambiguity between pathologists. Using PECNet allows the user to give the size of the HPF as an input. Comparing the PECNet PEC score (using a HPF size of 0.3mm$^2$) and the score from the CEGIR dataset resulted in a remarkable agreement between the machine and the human score and classification (Fig. 5). Translating the score agreement to the decision making depends on the threshold. When taking different thresholds and comparing the performance in the ROC space, the optimal accuracy is obtained with a threshold of 14.5/HPF; that is, for integer values, the best threshold is larger-equal than 15. This result highlights the lack of bias in PECNet's counting as this is the same threshold recommended by field experts[1].

Our work highlights the importance of multi-labeling capacity on small features and the ability to deploy the network on a whole slide image rapidly. These findings are a significant step towards automated diagnosis of EoE using digital microscopy and have implications for analyzing other biopsy-based disease diagnoses.

## Acknowledgments

Y.S. was supported by Israel Science Foundation #1619/20, Rappaport Foundation, and the Prince Center for Neurodegenerative Disorders of the Brain 3828931. CEGIR (U54 AI117804) is part of the Rare Disease Clinical Research Network (RDCRN), an initiative of the Office of Rare Diseases Research (ORDR), NCATS, and is funded through collaboration between NIAID, NIDDK, and NCATS. CEGIR is also supported by patient advocacy groups including American Partnership for Eosinophilic Disorders (APFED), Campaign Urging Research for Eosinophilic Diseases (CURED), and Eosinophilic Family Coalition (EFC). As a member of the RDCRN, CEGIR is also supported by its Data Management and Coordinating Center (DMCC) (U2CTR002818). This work is also supported by NIH R01 AI045898-21, the CURED Foundation, and Dave and Denise Bunning Sunshine Foundation. The authors would like to thank Yael Abuhatsera for technical support and valuable discussions.

**References**

1    Dellon ES, Liacouras CA, Molina-Infante J, *et al.* Updated International Consensus Diagnostic Criteria for Eosinophilic Esophagitis. In: Gastroenterology. W.B. Saunders, 2018: 1022-1033.e10.

2    Miehlke S. Clinical features of Eosinophilic esophagitis in children and adults. *Best Pract Res Clin Gastroenterol* 2015; 29: 739–48.

3    O'Shea KM, Aceves SS, Dellon ES, *et al.* Pathophysiology of Eosinophilic Esophagitis. *Gastroenterology* 2018; 154: 333–45.

4    Dellon ES. Eosinophilic esophagitis: Diagnostic tests and criteria. Curr. Opin. Gastroenterol. 2012. DOI:10.1097/MOG.0b013e328352b5ef.

5    Dellon ES, Fritchie KJ, Rubinas TC, Woosley JT, Shaheen NJ. Inter- and intraobserver reliability and validation of a new method for determination of eosinophil counts in patients with esophageal eosinophilia. *Dig Dis Sci* 2010; 55: 1940–9.

6    Stucke EM, Clarridge KE, Collins MH, Henderson CJ, Martin LJ, Rothenberg ME. Value of an Additional Review for Eosinophil Quantification in Esophageal Biopsies. *J Pediatr Gastroenterol Nutr* 2015; 61: 65–8.

7    Janowczyk A, Madabhushi A. Deep learning for digital pathology image analysis: A comprehensive tutorial with selected use cases. *J Pathol Inform* 2016; **7**.

8    Hart SN, Flotte W, Norgan AP, *et al.* Classification of melanocytic lesions in selected and whole-slide images via convolutional neural networks. *J Pathol Inform* 2019. DOI:10.4103/jpi.jpi_32_18.

9    Azevedo Tosta TA, Neves LA, do Nascimento MZ. Segmentation methods of H&E-stained histological images of lymphoma: A review. Informatics Med. Unlocked. 2017. DOI:10.1016/j.imu.2017.05.009.

10    Cui Y, Zhang G, Liu Z, Xiong Z, Hu J. A deep learning algorithm for one-step contour aware nuclei segmentation of histopathology images. *Med Biol Eng Comput* 2019. DOI:10.1007/s11517-019-02008-8.

11    Wang J, MacKenzie JD, Ramachandran R, Chen DZ. A deep learning approach for semantic segmentation in histology tissue images. In: Ourselin S, Joskowicz L, Sabuncu MR, Unal G, Wells W, eds. International Conference on Medical Image Computing and Computer-Assisted Intervention. Springer; Cham: Springer International Publishing, 2016: 176–84.

12    Kovalev V, Kalinovsky A, Liauchuk V. Deep Learning in Big Image Data: Histology Image Classification for Breast Cancer Diagnosis Protein docking by deep neural networks View project UAV: back to base problem View project. In: Int. Conference on BIG DATA and Advanced Analytics. 2016: 15–7.

13    Fakoor R, Nazi A, Huber M. Using deep learning to enhance cancer diagnosis and classification. In: The 30th International Conference on Machine Learning (ICML 2013),WHEALTH workshop. 2013. https://www.researchgate.net/publication/281857285.

14    Geras KJ, Wolfson S, Shen Y, Kim SG, Moy L, Cho K. High-Resolution Breast Cancer Screening with Multi-View Deep Convolutional Neural Networks. *Comput Res Repos*

2018; 1703.07047. http://arxiv.org/abs/1703.07047.

15    Djuric U, Zadeh G, Aldape K, Diamandis P. Precision histology: how deep learning is poised to revitalize histomorphology for personalized cancer care. *npj Precis Oncol* 2017; 1: 22.

16    Shen D, Wu G, Suk H-I. Deep Learning in Medical Image Analysis. *Annu Rev Biomed Eng* 2017; : Vol. 19:221-248.

17    He K, Gkioxari G, Dollár P, Girshick R. Mask R-CNN. *IEEE Trans Pattern Anal Mach Intell* 2020. DOI:10.1109/TPAMI.2018.2844175.

18    Johnson JW. Adapting Mask-RCNN for automatic nucleus segmentation. arXiv. 2018. DOI:10.1007/978-3-030-17798-0.

19    Chen LC, Papandreou G, Kokkinos I, Murphy K, Yuille AL. DeepLab: Semantic Image Segmentation with Deep Convolutional Nets, Atrous Convolution, and Fully Connected CRFs. *IEEE Trans Pattern Anal Mach Intell* 2018. DOI:10.1109/TPAMI.2017.2699184.

20    Chen LC, Papandreou G, Schroff F, Adam H. Rethinking atrous convolution for semantic image segmentation. arXiv. 2017.

21    Luc P, Couprie C, Chintala S, Verbeek J. Semantic Segmentation using Adversarial Networks. 2016.

22    Ronneberger O, Fischer P, Brox T. U-net: Convolutional networks for biomedical image segmentation. In: Lecture Notes in Computer Science (including subseries Lecture Notes in Artificial Intelligence and Lecture Notes in Bioinformatics). 2015. DOI:10.1007/978-3-319-24574-4_28.

23    Zhou Z, Siddiquee MMR, Tajbakhsh N, Liang J. UNet++: A nested U-Net architecture for medical image segmentation. arXiv. 2018.

24    Czyzewski T, Daniel N, Rochman M, Caldwell JM, Osswald GA, Margaret H. Machine learning approach for biopsy-based identification of eosinophilic esophagitis reveals importance of global features. ; : 1–5.

25    Gupta SK, Falk GW, Aceves SS, *et al.* Consortium of Eosinophilic Gastrointestinal Disease Researchers: Advancing the Field of Eosinophilic GI Disorders Through Collaboration. Gastroenterology. 2019. DOI:10.1053/j.gastro.2018.10.057.

26    Paszke A, Gross S, Massa F, *et al.* PyTorch: An imperative style, high-performance deep learning library. *Adv Neural Inf Process Syst* 2019; 32.12